\documentclass{article}

\usepackage[nonatbib,final]{nips_2017}

\usepackage[utf8]{inputenc} 
\usepackage[T1]{fontenc}    
\usepackage{hyperref}       
\usepackage{url}            
\usepackage{booktabs}       
\usepackage{amsfonts}       
\usepackage{nicefrac}       
\usepackage{microtype}      

\usepackage{graphicx}
\usepackage[square, numbers]{natbib}
\title{Transfer Learning with Human Corneal Tissues: \\ An Analysis of Optimal Cut-Off Layer}

\author{
  Nadezhda Prodanova\thanks{Institute for Automation and Applied Informatics (IAI), Karlsruhe Institute of Technology (KIT)} \\
  IAI, KIT \\
  Karlsruhe, Germany \\
  \And
  Johannes Stegmaier\\
  Institute of Imaging and Computer Vision\\
  RWTH Aachen University\\
  Aachen, Germany
  \And
  Stephan Allgeier\footnotemark[1]\\
  IAI, KIT \\
  Karlsruhe, Germany \\
  \And
  Sebastian Bohn\\
  Department of Ophthalmology\\ 
  Rostock University Medical Center\\
  Rostock, Germany\\
  \And
  Oliver Stachs\\
  Department of Ophthalmology\\ 
  Rostock University Medical Center\\
  Rostock, Germany\\
  \And
  Bernd Köhler\footnotemark[1]\\
  IAI, KIT \\
  Karlsruhe, Germany \\
  \And
  Ralf Mikut\footnotemark[1]\\
  IAI, KIT \\
  Karlsruhe, Germany \\
  \And
  Andreas Bartschat\footnotemark[1]\\
  IAI, KIT \\
  Karlsruhe, Germany \\
  \texttt{andreas.bartschat@kit.edu}
}

\begin{document}

\maketitle

\begin{abstract}
Transfer learning is a powerful tool to adapt trained neural networks to new tasks. Depending on the similarity of the original task to the new task, the selection of the cut-off layer is critical. For medical applications like tissue classification, the last layers of an object classification network might not be optimal. We found that on real data of human corneal tissues the best feature representation can be found in the middle layers of the Inception-v3 and in the rear layers of the VGG-19 architecture.
\end{abstract}

\section{Introduction}
Deep neural networks (DNN) trained on large datasets like ImageNet \cite{Russakovsky15} provide high performance in transfer learning tasks for medical applications even for small training datasets \cite{Huh2016, Litjens17}. In general, the trained  basis networks can be used either for feature extraction or fine-tuned to perform new tasks.

If the network is used for feature extraction, usually the last fully connected layers are cut off and replaced by new layers or a classifier like a support vector machine, which is trained on the output of the network. However, the layer with the most discriminant features should be used to get the best performance.

Analysis of DNN shows that the first layer performs a class agnostic feature extraction (\textit{e.g.} Gabor or RGB related filters) and the last layer learns highly class-specific representations. The intermediate layers are expected to create different levels of abstractions to connect these two extreme cases \cite{Azizpour15, Yosinski14}. Depending on the similarity of the new task to the original task, the level of abstraction must be chosen and therefore the layer where the network is cut off. 

Especially for medical applications, where transfer learning can be used to classify homogeneous areas of  specific tissues, the choice of the cut-off layer is an important parameter. Since the original network is trained for object classification, not only the last layer might be involved in the creation of class-specific features. Hence, the more generic middle layers might provide a preferable feature representation for the new task. For the training of the new layers, no regularization or data augmentation is applied.

This abstract presents a study on transfer learning for corneal tissue classification with tissue layers of epithelium, sub-basal nerve plexus (SNP) and stroma images. The images are acquired by a laser scanning confocal microscope \cite{Allgeier14} previously analyzed with a Bag-of-Words approach \cite{Bartschat16FBV}. We analyze the impact of the cut-off layer with the VGG-19 \cite{Arge2015} and the Inception-v3 \cite{Szegedy2015} architectures, both trained on the ImageNet dataset.

\section{Methods}
\begin{figure}[h]
  \centering
  \includegraphics[width=.8\linewidth]{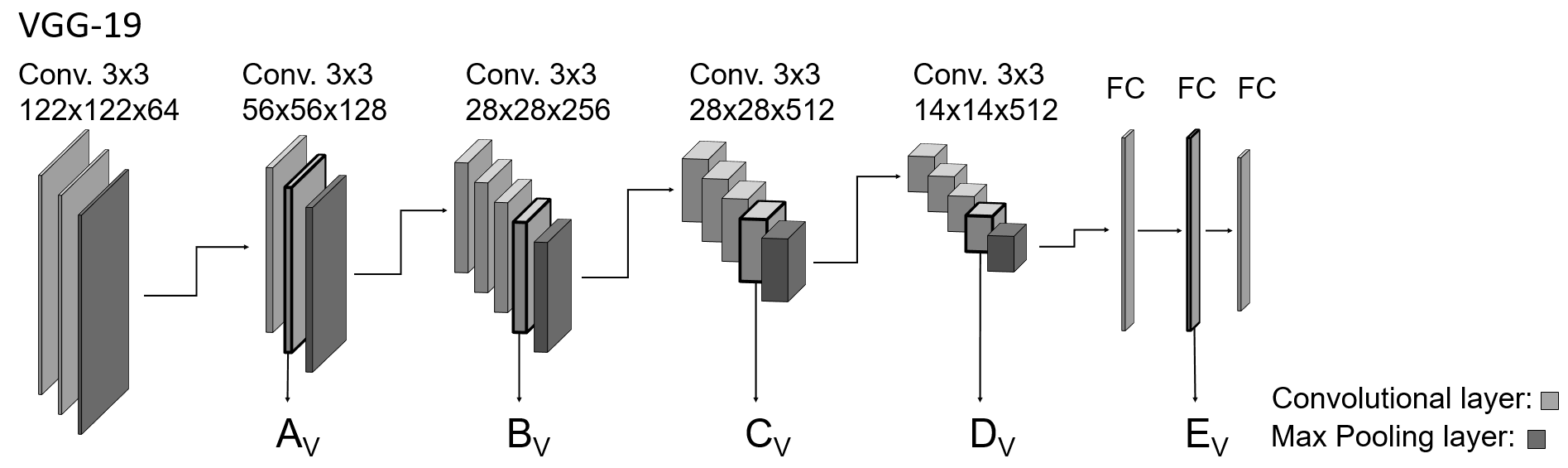}
  \caption{VGG-19 architecture with marked cut-off layers used for transfer learning.}
  \label{fig:vgg}
\end{figure}

For the analysis of the optimal cut-off layer of the networks, multiple layers are analyzed. For the VGG-19 architecture, we investigated four layers ($A_V$-$D_V$) in front of the max-pooling layer as well as a fully connected layer at the end ($E_V$) as shown in Figure \ref{fig:vgg}. The Inception-v3 model combines two blocks of convolutions followed by max-pooling, as well as three blocks of inception modules \cite{Szegedy2015}. We investigated four layers, the first layer ($A_I$) is the last convolution layer in front of the inception modules and the next three layers ($B_I$-$D_I$) are taken at the end of the inception module blocks.  

Based on the size of the output tensor, we perform a downsampling with a max-pooling layer to represent every input image with less than $8000$ values. After flattening this tensor, a fully connected layer is trained with a softmax approach for the new classification task.

The dataset with $5597$ images from eleven healthy human volunteers is divided into a training, an evaluation and a test set. Since the images from a single volunteer might have overlapping regions, this separation was done manually to avoid similar pictures in the different datasets. The training set consists of about $83\%$ of the data, the evaluation and the test set consist of $7\%$ and $10\%$.

\section{Results}
\begin{figure}[h]
  \centering
  \includegraphics[width=.49\linewidth]{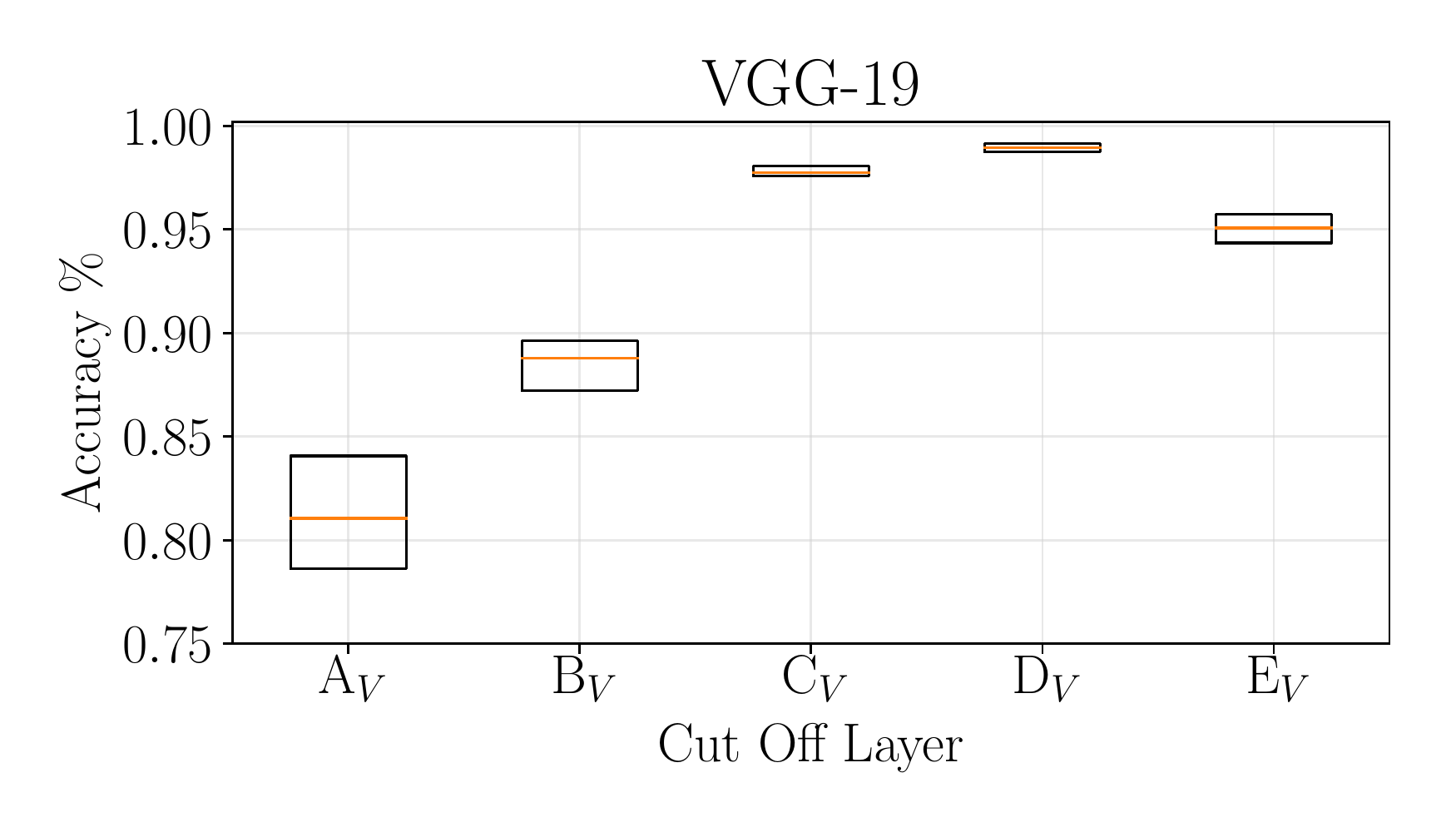}
  \includegraphics[width=.49\linewidth]{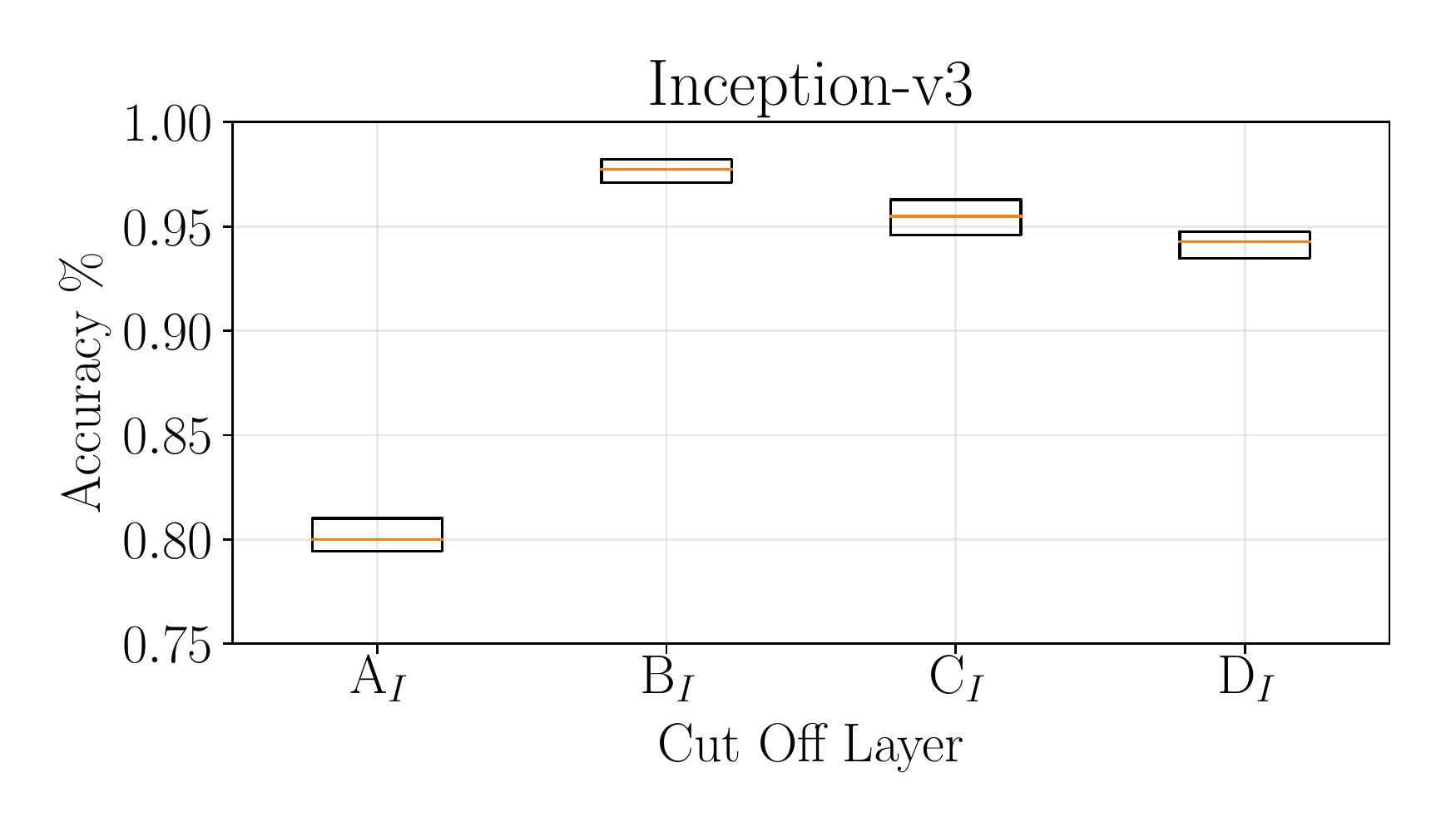}
  \caption{Classification accuracy for different cut-off layers with the VGG-19 (left) and the Inception-v3 architecture (right).}
  \label{fig:results}
\end{figure}

Figure \ref{fig:results} shows the achieved accuracy with different cut-off layers of VGG-19 and Inception-v3 for ten runs for each layer. For the VGG-19 model, the cut off at layer $D_V$ leads to the best mean accuracy of $98.9\%$ followed by layer $C_V$. Thus, using the fully connected layer at the end reduces the transfer learning performance for the given task. This effect is even more notable in the Inception-v3 model. Here, layer $B_I$ leads to the best result with a mean accuracy of $97.1\%$ while the accuracy is decreasing with each additional layer. Meaning the feature representation of the first inception module block is best for to the investigated task and the embedding gets worse with further inception blocks.
Hence, for the given task the Inception-v3 architecture can be reduced to less than $10\%$ of its parameters with better accuracy and a significant reduction of computation time.

\section{Conclusion}
We investigated the classification performance for transfer learning with the Inception-v3 and VGG-19 architectures for the classification of corneal tissues. Both architectures were trained on the ImageNet dataset. Since the medical images do not contain single objects but show a homogeneous tissue over the whole image, the new classification task differs significantly from the original.

We analyzed the performance of different layers for transfer learning and found that the optimal feature representation for medical images can be found in the middle of the investigated networks, since the last layers are too specialized for the original task. On the other hand, a complete retraining including the first layer is inefficient for the relatively small dataset. 

Further investigations of the feature representation might be useful to find metrics correlated to the classification accuracy to select the optimal cut-off layer fast and without the training of a classifier. Moreover, the accuracy might be increased by using combinations of different layers or different dimensionality reduction techniques as important factors of the classification performance.

\bibliography{bib}

\end{document}